\documentclass[10pt,twocolumn,letterpaper]{article}

\usepackage{iccv}
\usepackage{times}
\usepackage{epsfig}
\usepackage{graphicx}
\usepackage{amsmath}
\usepackage{amssymb}
\newcommand\blfootnote[1]{%
  \begingroup
  \renewcommand\thefootnote{}\footnote{#1}%
  \addtocounter{footnote}{-1}%
  \endgroup
}

\usepackage[pagebackref=true,breaklinks=true,letterpaper=true,colorlinks,bookmarks=false]{hyperref}

\iccvfinalcopy 

\def\iccvPaperID{3392} 

\ificcvfinal\fi
\begin{document}

\title{R-FCN-3000 at 30fps: Decoupling Detection and Classification}

\author{Bharat Singh*$^{1}$~~~~~~~~Hengduo Li*$^{2}$~~~~~~~~Abhishek Sharma$^{3}$~~~~~~~~Larry S. Davis$^{1}$\\
University of Maryland, College Park $^{1}$~~~Fudan University$^{2}$~~~Gobasco AI Labs$^{3}$\\
{\tt\small \{bharat,lsd\}@cs.umd.edu~~~~lihd14@fudan.edu.cn~~~~abhisharayiya@gmail.com}}

\maketitle
\blfootnote{*Equal Contribution. Work done during H. Li's internship at UMD.}

\begin{abstract}
We present R-FCN-3000, a large-scale real-time object detector in which objectness detection and classification are decoupled. To obtain the detection score for an RoI, we multiply the objectness score with the fine-grained classification score. Our approach is a modification of the R-FCN architecture in which position-sensitive filters are shared across different object classes for performing localization. For fine-grained classification, these position-sensitive filters are not needed. R-FCN-3000 obtains an mAP of 34.9\% on the ImageNet detection dataset and outperforms YOLO-9000 by 18\% while processing 30 images per second. We also show that the objectness learned by R-FCN-3000 generalizes to novel classes  and the performance increases with the number of training object classes - supporting the hypothesis that it is possible to learn a universal objectness detector. Code will be made available.
\end{abstract}

\section{Introduction}

With the advent of Deep CNNs \cite{hinton2012deep,krizhevsky2012imagenet}, object-detection has witnessed a quantum leap in the performance on benchmark datasets. It is due to the powerful feature learning capabilities of deep CNN architectures. Within the last five years, the mAP scores on PASCAL \cite{everingham2010pascal} and COCO \cite{lin2014microsoft} have improved from 33\% to 88\% and 37\% to 73\% (at 50\% overlap), respectively. While there have been massive improvements on standard benchmarks with tens of classes \cite{girshick2014rich,girshick2015fast,ren2015faster,dai2017deformable,he2017mask}, little progress has been made towards real-life object detection that requires real-time detection of thousands of classes. Some recent efforts \cite{redmon2016yolo9000,hoffman2014lsda} in this direction have led to large-scale detection systems, but at the cost of accuracy. We propose a solution to the large-scale object detection problem that outperforms YOLO-9000 \cite{redmon2016yolo9000} by 18\% and can process 30 images per second while detecting 3000 classes, referred to as R-FCN-3000.

R-FCN-3000 is a result of systematic modifications to some of the recent object-detection architectures \cite{dai2017deformable, dai2016r, lin2017focal, liu2016ssd, redmon2016you} to afford real-time large-scale object detection. Recently proposed fully convolutional class of detectors \cite{dai2017deformable, dai2016r, lin2017focal, liu2016ssd, redmon2016you} compute per-class objectness score for a given image. They have shown impressive accuracy within limited computational budgets. Although fully-convolutional representations provide an efficient \cite{huang2016speed} solution for tasks like object detection \cite{dai2017deformable}, instance segmentation \cite{Li_2017_CVPR}, tracking \cite{feichtenhofer2017detect},  relationship detection \cite{zhang2017ppr} etc., they require class-specific sets of filters for each class that prohibits their application for large number of classes. For example, R-FCN \cite{dai2016r}/ Deformable-R-FCN \cite{dai2017deformable} requires 49/197 position-specific filters for each class. Retina-Net \cite{lin2017focal} requires 9 filters for each class for each convolutional feature map. Therefore, such architectures would need hundreds of thousands of filters for detecting 3000 classes, which will make them extremely slow for practical purposes.
 \begin{figure}
    \center
    \includegraphics[width=0.99\linewidth]{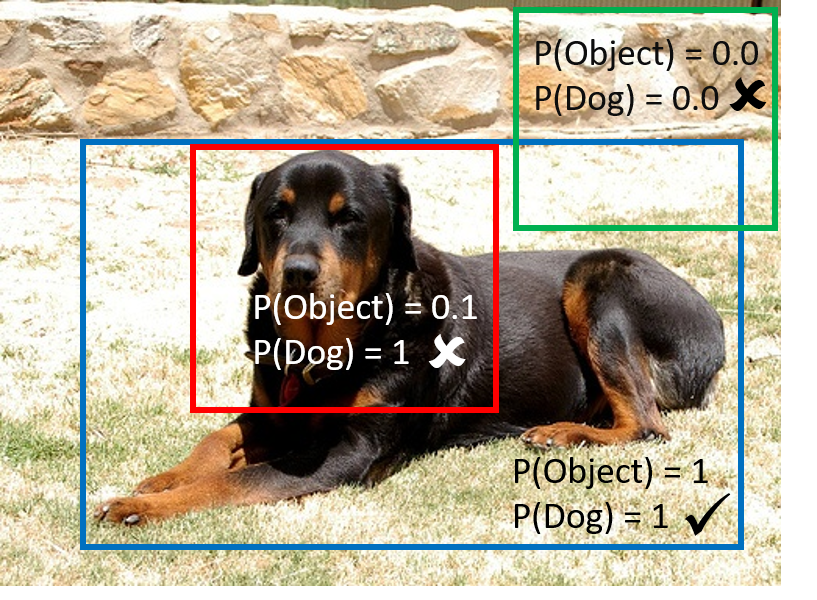}
    \caption{We propose to decouple classification and localization by independently predicting objectness and classification scores. These scores are multiplied to obtain a detector.}
    \label{fig:dog}
\end{figure}

The key insight behind the proposed R-FCN-3000 architecture is to decouple {\em objectness detection } and classification of the detected object so that the computational requirements for localization remain constant as the number of classes increases - see Fig. \ref{fig:dog}. We leverage the fact that many object categories are visually similar and share parts. For example - different breeds of dogs all have common body parts; therefore, learning a different set of filters for detecting each breed is overkill. So, R-FCN-3000 performs object detection (with position-sensitive filters) for a fixed number of {\em super-classes} followed by fine-grained classification (without position-sensitive filters) within each super-class. The super-classes are obtained by clustering the deep semantic features of images (2048 dimensional features of ResNet-101 in this case); therefore, we do not require a semantic hierarchy. The fine-grained class probability at a given location is obtained by multiplying the super-class probability with the classification probability of the fine-grained category within the super-class. 

In order to study the effect of using super-classes instead of individual object categories, we varied the number of super-classes from 1 to 100 and evaluated the performance on the ImageNet detection dataset. Surprisingly, the detector performs well even with one super-class! This observation indicates that position-sensitive filters can potentially learn to detect universal objectness. It also reaffirms a well-researched concept from the past \cite{alexe2012measuring,arbelaez2011contour, uijlings2013selective} that objectness is a generic concept and a universal objectness detector can be learned. Thus, for performing object detection, it suffices to multiply the objectness score of an RoI with the classifiation probability for a given class. This results in a fast detector for thousands of classes, as per-class position sensitive filters are no longer needed. On the PASCAL-VOC dataset, with only our objectness based detector, we observe a 1.5\% drop in mAP compared to the deformable R-FCN \cite{dai2017deformable} detector with class-specific filters for all 20 object classes. R-FCN-3000, trained for 3000 classes, obtains an 18\% improvement in mAP over the current state-of-the-art large scale object detector (YOLO-9000) on the ImageNet detection dataset. Finally, we also evaluate the generalizability of our objectness detector on {\em unseen} classes (a zero-shot setting for localization) and observe that the generalization error decreases as we train the objectness detector on larger numbers of classes.

\section{Related Work}
Large scale localization using deep convolutional networks was first performed in \cite{sermanet2013overfeat, simonyan2014very} which used regression for predicting the location of bounding boxes. Later, RPN \cite{ren2015faster} was used for localization in ImageNet classification \cite{he2016deep}. However, no evaluations were performed to determine if these networks generalize when applied on detection datasets without specifically training on them. Weakly-supervised detection has been a major focus over the past few years for solving large-scale object detection. In \cite{hoffman2014lsda}, knowledge of detectors trained with bounding boxes was transferred to classes for which no bounding boxes are available. The assumption is that it is possible to train object detectors on a fixed number of classes. For a class for which supervision is not available, transformations are learned to adapt the classifier to a detector. Multiple-instance learning based approaches have also been proposed which can leverage weakly supervised data for adapting classifiers to detectors \cite{hoffman2016large}. Recently, YOLO-9000 \cite{redmon2016yolo9000} jointly trains on classification and detection data. When it sees a classification image, classification loss is back-propagated on the bounding box which has the highest probability. It assumes that the predicted box is the ground truth box and uses the difference between other anchors and the predicted box as the objectness loss. YOLO-9000 is fast, as it uses a lightweight network and uses 3 filters per class for performing localization. For performing good localization, just 3 priors are not sufficient.

For classifying and localizing a large number of classes, some methods leverage the fact that parts can be shared across objects categories \cite{novotny2016have,salakhutdinov2011learning,torralba2004sharing,ott2011shared}. Sharing filters for object parts reduces model complexity and also reduces the amount of training data required for learning part-based filters. Even in traditional methods, it has been shown that when filters are shared, they are more generic \cite{torralba2004sharing}. However, current detectors like Deformable-R-FCN \cite{dai2017deformable}, R-FCN \cite{dai2016r}, RetinaNet \cite{lin2017focal} do not share filters (in the final classification layer) across object categories: because of this, inference is slow when they are applied on thousands of categories. Taking motivation from prior work on sharing filters across object categories, we propose an architecture where filters can be shared across some object categories for large scale object detection.

The extreme version of sharing parts is objectness, where we assume that all objects have something in common. Early in this decade (if not before), it was proposed that objectness is a generic concept and it was demonstrated that only a very few category agnostic proposals were sufficient to obtain high recall \cite{uijlings2013selective,carreira2010constrained,arbelaez2011contour,alexe2012measuring}. With a bag-of-words feature-representation \cite{lazebnik2006beyond} for these proposals, better performance was shown compared to a sliding-window based part-based-model \cite{felzenszwalb2010object} for object detection. R-CNN \cite{girshick2014rich} used the same proposals for object detection but also applied per-class bounding-box regression to refine the location of these proposals. Subsequently, it was observed that per-class regression was not necessary and a class-agnostic regression step is sufficient to refine the proposal position \cite{dai2016r}. Therefore, if the regression step is class agnostic, and it is possible to obtain a reasonable objectness score, a simple classification layer should be sufficient to perform detection. We can simply multiply the objectness probability with the classification probability to make a detector! Therefore, in the extreme case, we set the number of super-classes to one and show that we can train a detector which obtains an mAP which is very close to state-of-the-art object detection architectures \cite{dai2016r}.

\section{Background}
This section provides a brief introduction of Deformable R-FCN \cite{dai2017deformable} which is used in R-FCN-3000. In R-FCN \cite{dai2016r}, {\em Atrous} convolution \cite{chen2016deeplab} is used in the conv5 layer to increase the resolution of the feature map while still utilizing the pre-trained weights from the ImageNet classification network. In Deformable-R-FCN \cite{dai2017deformable}, the atrous convolution is replaced by a deformable convolution structure in which a separate branch predicts offsets for each pixel in the feature map, and the convolution kernel is applied after the offsets have been applied to the feature-map. A region proposal network (RPN) is used for generating object proposals, which is a two layer CNN on top of the conv4 features. Efficiently implemented local convolutions, referred to as position sensitive filters, are used to classify these proposals.

\section{Large Scale Fully-Convolutional Detector}
This section describes the process of training a large-scale object detector. We first explain the training data requirements followed by discussions of some of the challenges involved in training such a system - design decisions for making training and inference efficient, appropriate loss functions for a large number of classes, mitigating the domain-shift which arises when training on classification data. 

\subsection{Weakly Supervised vs. Supervised?}
Obtaining an annotated dataset of thousands of classes is a major challenge for large scale detection. Ideally, a system that can learn to detect object instances using partial image level tags (class labels) for the objects present in training images would be preferable because large-scale training data is readily available on the internet in this format. Since the setting with partial annotations is very challenging, it is commonly assumed that labels are available for all the objects present in the image. This is referred to as the {\em weakly supervised} setting. Unfortunately, explicit boundaries of objects or atleast bounding-boxes are required as supervision signal for training accurate object detectors. This is the {\em supervised} setting. The performance gap between supervised and weakly supervised detectors is large - even 2015 object detectors \cite{he2016deep} were better by 40\% on the PASCAL VOC 2007 dataset compared to recent weakly supervised detectors \cite{diba2016weakly}. This gap is a direct result of insufficient learning signal coming from weak supervision and can be further explained with the help of an example. For classifying a dog among 1000 categories, only body texture or facial features of a dog may be sufficient and the network need not learn the visual properties of its tail or legs for correct classification. Therefore, it may never learn that legs or tail are parts of the dog category, which are essential to obtain accurate boundaries. 

On one hand, the huge cost of annotating bounding boxes for thousands of classes under settings similar to popular detection datasets such as PASCAL or COCO makes it prohibitively expensive to collect and annotate a large-scale detection dataset. On the other hand, the poor performance of weakly supervised detectors impedes their deployment in real-life applications. Therefore, we ask - is there a middle ground that can alleviate the cost of annotation while yielding accurate detectors? Fortunately, the ImageNet database contains around 1-2 objects per image; therefore, the cost of annotating the bounding boxes for the objects is only a few seconds compared to several minutes in COCO \cite{lin2014microsoft}. It is because of this reason that the bounding boxes were also collected while annotating ImageNet! A potential downside of using ImageNet for training object detectors is the loss of variation in scale and context around objects available in detection datasets, but we do have access to the bounding-boxes of the objects. Therefore, a natural question to ask is, how would an object detector perform on ``detection'' datasets if it were trained on classification datasets with bounding-box supervision? We show that careful design choices with respect to the CNN architecture, loss function and training protocol can yield a large-scale detector trained on the ImageNet classification set with significantly better accuracy compared to weakly supervised detectors.

\begin{figure*}
    \center
    \includegraphics[width=1.00\linewidth]{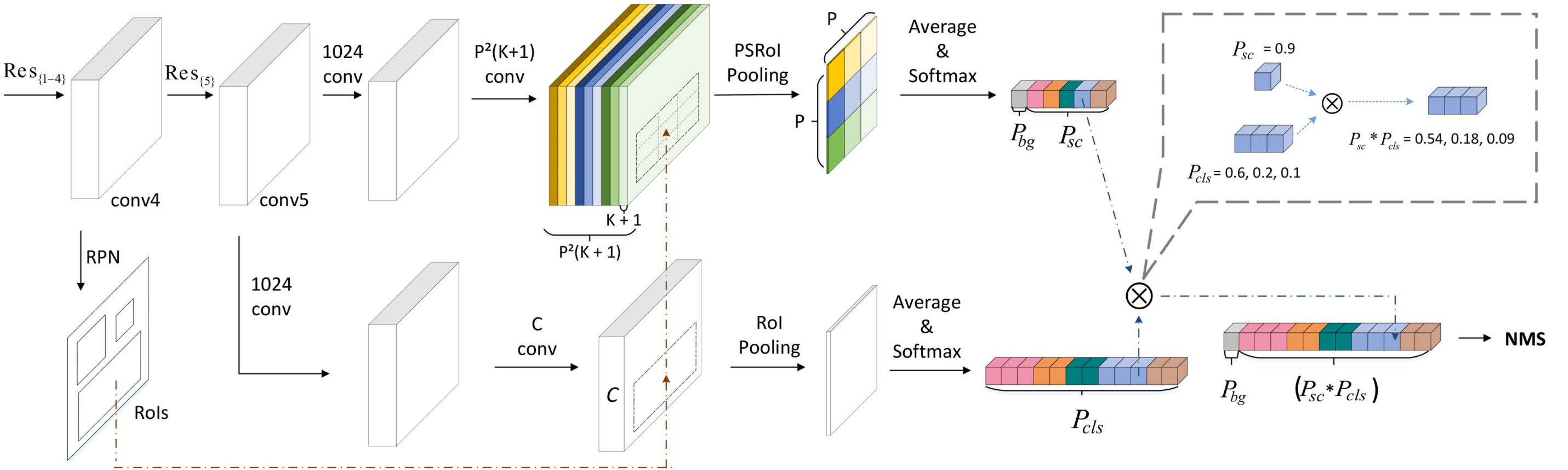}
    \caption{R-FCN-3000 first generates region proposals which are provided as input to a super-class detection branch (like R-FCN) which jointly predicts the detection scores for each super-class (sc). A class-agnostic bounding-box regression step refines the position of each RoI (not shown). To obtain the semantic class, we do not use position-sensitive filters but predict per class scores in a fully convolutional fashion. Finally, we average pool the per-class scores inside the RoI to get the classification probability. The classification probability is multiplied with the super-class detection probability for detecting 3000 classes. When K is 1, the super-class detector predicts objectness.}
    \label{fig:arch}
\end{figure*}

\subsection{Super-class Discovery} \label{sec:superclass}

Fully convolutional object detectors learn class-specific filters based on scale \& aspect-ratio \cite{lin2017focal} or in the form of position sensitive filters \cite{dai2016r,dai2017deformable} for each class. Therefore, when the number of classes become large, it becomes computationally in-feasible to apply these detectors. Hence, we ask is it necessary to have sets of filters for each class or can they be shared across visually similar classes? In the extreme case - can detection be performed using just a foreground/background detector and a classification network? To obtain visually similar sets of objects for which position-sensitive filters can be shared, objects should have similar visual appearances. We obtain the $j^{th}$ object-class representation, $x_{j}$, by taking the average of 2048-dimensional feature-vectors ($x^i_j$), from the final layer of ResNet-101, for the all the samples belonging to the $j^{th}$ object-class in the ImageNet classification dataset (validation set). Super-classes are then obtained by applying K-means clustering on $\{ x_j : j \in \{1, 2, \dots C\}\}$, where $C$ is the number of object-classes, to obtain $K$ super-class clusters.

\subsection{Architecture} \label{sec:arch}
First, RPN is used for generating proposals, as in \cite{dai2017deformable}. Let the set of individual object-classes the detector is being trained on be $\mathcal{C}, |\mathcal{C}| = C$, and the set of super-classes (SC) be $\mathcal{K}, |\mathcal{K}| = K$. For each super-class $k$, suppose we have $P \times P$ position-sensitive filters, as shown in Fig \ref{fig:arch}. On the conv5 feature, we first apply two independent convolution layers as in R-FCN for obtaining detection scores and bounding-box regression offsets. On each of these branches, after a non-linearity function, we apply position sensitive filters for classification and bounding-box regression. Since we have $K$ super-classes and $P \times P$ filters per super-class, there are $(K + 1) \times P \times P$ filters in the classification branch (1 more for background) and $P \times P$ filters in the bounding-box regression branch as this branch is class-agnostic. After performing position-sensitive RoI pooling and averaging the predictions in each bin, we obtain predictions of the network for classification and localization. To get the super-class probability, softmax function over $K$ super-classes is used and predictions from the localization branch are directly added to get the final position of the detection. These two branches help detect the super-classes which are represented by each cluster $k$. For obtaining fine-grained class information, we employ a two layer CNN on the conv5 feature map, as shown in Fig . \ref{fig:arch}. A softmax function is used on the output of this layer for obtaining the final class probability. The detection and classification probabilities are multiplied to obtain the final detection score for each object-class. This architecture is shown in Fig. \ref{fig:arch}. Even though there are other challenges such as entailment, cover, equivalence etc. \cite{maccartney2009extended,deng2014large} which are not correctly modelled with the softmax function, the Top-1 accuracy even on the ImageNet-5000 classification dataset is greater than 67\% \cite{Chen2017}. So, we believe these are not the bottlenecks for detecting a few thousand classes. 

\subsection{Label Assignment} \label{sec:lab-assign}
Labels are assigned exactly the same way as fast-RCNN \cite{girshick2015fast} for the $K$ super-classes on which detection is performed. Let $C$ be the total number of object-classes and let $k_i$ and $c_j$ denote the $i^{th}$ super-class and $j^{th}$ sub-class in $k_i$, then $k_i = \{c_1, c_2, ..., c_j\}$ and $\sum_{i=1}^{K} |k_i| = C$. For detecting super-class $k_{i}$, an RoI is assigned as positive for super-class $k_i$ if it has an intersection over union (IoU) greater than 0.5 with any of the ground truth boxes in $k_i$, otherwise it is marked as background (label for background class $K+1$ is set to one). For the classification branch (to get the final 3000 classes), only positive RoIs are used for training, i.e. only those which have an IoU greater than 0.5 with a ground truth bounding box. The number of labels for classification is $C$ instead of $K+1$ in detection.

\subsection{Loss Function}\label{sec:loss}
For training the detector, we use online hard example mining (OHEM) \cite{shrivastava2016training} as done in \cite{dai2017deformable} and smooth L1 loss for bounding box localization \cite{girshick2015fast}. For fine-grained classification we only use a softmax loss function over $C$ object-classes for classifying the positive bounding boxes. Since the number of positive RoIs are typically small compared to the number of proposals, the loss from this branch is weighted by a factor of 0.05, so that these gradients do not dominate network training. This is important as we train RPN layers, R-FCN classification and localization layers, and fine-grained layers together in a multi-task fashion, so balancing the loss from each branch is important.



\section{Experiments}
In this section, we describe the implementation details of the proposed large-scale object detector and compare against some of the weakly supervised large-scale object detectors in terms of speed and accuracy.

\subsection{Training Data}
We train on the ImageNet classification dataset which contains bounding boxes for 3,130 classes. Each class contains at least 100 images in the training set. In the complete dataset, there are 1.2 million images. The detection test set (ILSVRC 2014) of ImageNet contains 194 classes out of the 3,130 classes present in the classification set. Therefore, we present our results on the 194 classes in our experiments (6 classes did not have bounding boxes in the ImageNet classification dataset which were present in the ImageNet detection test set). We also perform experiments on the PASCAL VOC 2007+2012 object detection dataset. The PASCAL dataset contains 20 object classes and 15,000 images for training. We evaluate our models on the VOC 2007 test set.

\begin{table}[t!]
    \centering
    \begin{tabular}{c|c|c}
        \hline
          Dataset (ImageNet) & Images & Object Instances \\
        \hline
         Detection & 400,000 & 764,910  \\
        \hline
         CLS 194 & 87,577 & 100,724  \\
        \hline
         CLS 500 & 121,450 & 141,801  \\
        \hline
         CLS 1000 & 191,463 & 223,222  \\
        \hline
         CLS 2000 & 403,398 & 462,795  \\
        \hline
         CLS 3000 & 925,327 & 1,061,647  \\
        \hline
    \end{tabular}
    \newline
    \caption{The number of images and object instances in the ImageNet Detection and different versions of our ImageNet classification (CLS) training set.}
    \label{tab:stats}
\end{table}

\subsection{Implementation Details}
For fast training and inference, we train on images of resolution (375x500), where the smaller side is at least 375 pixels and the larger side is a maximum of 500 pixels. Three anchor scales of (64,128,256) pixels are used. At each anchor scale, there are 3 aspect ratios of (1:2), (1:1) and (2:1) for the anchor boxes, hence there are a total of 9 anchors in RPN. We train for 7 epochs. A warm-up learning rate of 0.00002 is used for first 1000 iterations and then it is increased  to 0.0002. The learning rate is dropped by a factor of 10 after 5.33 epochs. Training is performed on 2 Nvidia P6000 GPUs. End-to-end training on 3,130 classes takes 2 weeks. When increasing the number of classes beyond 194, we first select classes with the least number of samples (each class still has at least 100 samples) from the classification set. This is done for accelerating our ablation experiments on 500, 1000 and 2000 classes. Statistics of the detection set and classification set with different numbers of classes are shown in Table \ref{tab:stats}. For faster training and inference, we do not use deformable position-sensitive RoI pooling and only use bi-linear interpolation. For our analysis, we first train a region proposal network on 3,130 classes separately and extract proposals on the training and test set. Then, deformable R-FCN is trained like fast-RCNN with different numbers of clusters and classes. This is also done to accelerate training. It takes 2 days to train on 1000 classes after making these changes. During training, horizontal flipping is used as a data augmentation technique. Multi-scale inference is performed at two scales, (375,500) and (750,1000) and predictions of the two scales are combined using NMS. In all our experiments, a ResNet-50 backbone network is used. On the PASCAL VOC dataset we train under the same settings as \cite{dai2017deformable}.
\begin{figure*}[!ht]
    \center
    \includegraphics[width=0.99\linewidth]{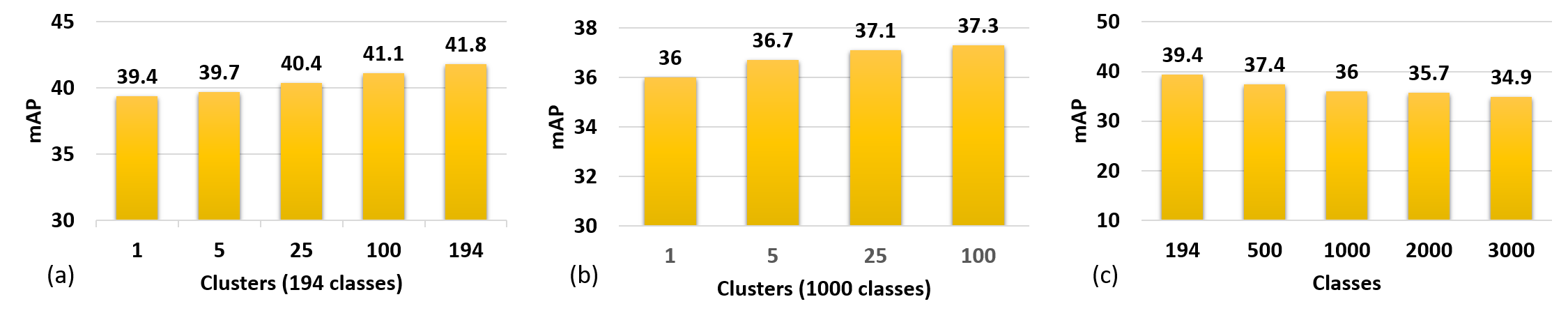}
    \caption{The mAP on the 194 classes in the ImageNet detection set is shown as we vary the number of clusters (super-classes). This is shown for 194 class and 1000 class detectors. We also plot the mAP for different number of classes for an objectness based detector.}
    \label{fig:numc}
\end{figure*}

\begin{table}[t!]
    \centering
    \begin{tabular}{c|c|c|c|c}
        \hline
         Method  & LSDA \cite{hoffman2014lsda} & SKT \cite{tang2016large} & KDT \cite{uijlings2013selective} & Ours  \\
        \hline
         mAP  & 18.1 & 20.0 & 34.3 & 43.3 \\
        \hline
    \end{tabular}
    \newline
    \caption{Comparison of our decoupled R-FCN trained on classification data with bounding-box supervision vs. weakly-supervised methods that use a knowledge transfer approach to exploit information from detectors pre-trained on 100 classes on the ImageNet detection set.}
    \label{tab:weak}
\end{table}

\subsection{Comparison with Weakly Supervised Detectors}
First, to calibrate our results with existing methods and to highlight the improvement by training on classification data with bounding-box supervision, we compare our method with knowledge transfer based weakly supervised methods. Methods like LSDA \cite{hoffman2014lsda} and Semantic Knowledge Transfer (SKT) \cite{tang2016large} assume that detectors for 100 classes (trained on the ImageNet detection dataset) are available and use semantic similarity between weakly supervised classes and strongly supervised classes to leverage information learned from pre-trained detectors. They evaluate on the remaining 100 classes in the ImageNet detection set. Contemporary work  \cite{uijlings2017revisiting} (KDT) also employs a knowledge transfer based approach, albeit with a modern Inception-ResNet-v2 based Faster-RCNN detector. Since these methods leverage classification data and also detection data for other classes, these can be considered as a very loose upper-bound on what a {\em true weakly-supervised detector}, which does not have any access to bounding boxes, would achieve. Our single scale ResNet-50 based model trained on 194 classes obtains an mAP of 40.5\% \footnote{2 classes did not have bounding box annotations in the ImageNet classification training set, so results are on 98 out of 100 classes} and after multi-scale testing (2 scales), we obtain an mAP of 43.3\%. So, with bounding-box supervision on classification data, we obtain a 9\% improvement. The performance of our detector can be further improved with a stronger backbone architecture, with deformable PSRoI layers etc. Note that we compare our method trained on 194 classes because a detector trained on larger number of classes is likely to perform worse if the new classes do not frequently occur in the test set. 

We also provide some statistics on the number of images and object instances in the ImageNet detection and ImageNet classification set in Table \ref{tab:stats}. Weakly supervised methods like LSDA \cite{hoffman2014lsda}, SKT \cite{tang2016large}, KDT \cite{uijlings2013selective} use detectors trained on 400,000 instances present in 200,000 images from the detection dataset. This acts as a prior which is used as a basis for adapting the classification network. We compare with these methods as we could not find methods with superior performance which are trained in a completely weakly supervised way on the ImageNet classification/detection dataset and are evaluated on the ImageNet detection set. 


\begin{table}[t!]
    \centering
    \begin{tabular}{c|c|c|c|c|c}
        \hline
          Clusters & 1 & 5 & 25 & 100 & 1000 \\
        \hline
          mAP & 36 & 36.7 & 37.1 & 37.3 & - \\
        \hline
         Time(ms) & 33 & 33 & 34 & 51 & 231 \\
        \hline
    \end{tabular}
    \newline
    \caption{The mAP scores for different number of clusters for the 1000 class detector and run-time(in milli-seconds)/image.}
    \label{tab:speed}
\end{table}
\subsection{Speed and Performance}
In Table \ref{tab:speed}, we present the speed accuracy trade-off when increasing the number of clusters for the 1000 class detector. The 100 class clustering based detector is 66\% slower than the objectness based detector. It was infeasible to train the original detector with 1000 classes, so we only present the run time for this detector. All the speed results are on a P6000 GPU. We also present results when we use different numbers of clusters during NMS. In this process, NMS is performed for a group of visually similar classes together, instead of each class separately. We use the same clustering based grouping of classes. The clusters used during NMS can be different from those which are used when grouping classes for R-FCN as this is only done for accelerating the post-processing step. We present the runtime for NMS (on GPU) for different numbers of clusters in Table \ref{tab:nms}. Note that 10 ms is 33\% of the runtime of our detector, and this is only for 1000 classes. Therefore, performing NMS on visually similar classes is a simple way to speed up inference without taking a significant hit in average precision. As mentioned in the title, our 3000 class detector can be applied to more than 30 images per second (on a resolution of 375x500 pixels, minimum side 375, maximum side 500) on an Nvidia P6000 GPU.

\begin{table}[t!]
    \centering
    \begin{tabular}{c|c|c|c|c|c}
        \hline
          Clusters & 20 & 50 & 100 & 200 & 1000 \\
        \hline
          mAP & 35.6 & 35.6 & 35.6 & 35.7 & 36.0 \\
        \hline
         Time(ms) & 1 & 1.5 & 1.8 & 2.6 & 10.1 \\
        \hline
    \end{tabular}
    \newline
    \caption{The mAP for different number of super-classes in NMS for the 1000 class objectness based detector and the NMS run-time (in milli-seconds).}
    \label{tab:nms}
\end{table}

\begin{figure*}[!t]
    \center
    \includegraphics[width=1.00\linewidth]{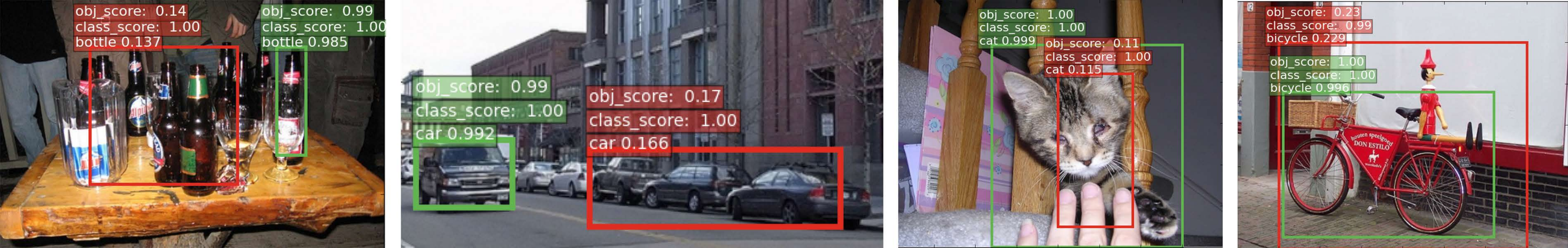}
    \caption{The objectness, classification and final detection scores against various transformations such as combinations of scaling and translation are shown. These scores are generated by forward propagating an ideal bounding-box RoI (in green) and a transformed bounding-box RoI (in red) through the R-FCN (objectness) and classification branch of the network. The selectiveness of the detector in terms of objectness is clearly visible against the various transformations that lead to poor detection.
    }
    \label{fig:pascal}
\end{figure*}

\section{Discussion}
In order to better understand the behaviour of the proposed object detection system, we evaluate it while varying the number of clusters and classes under different training and testing dataset conditions. Lastly, we also conduct experiments with unseen classes during training to assess the generalization performance of the proposed detector beyond the training classes.
\subsection{Impact of Number of Classes and Clusters}
\begin{figure*}[ht!]
    \center
    \includegraphics[width=1.00\linewidth]{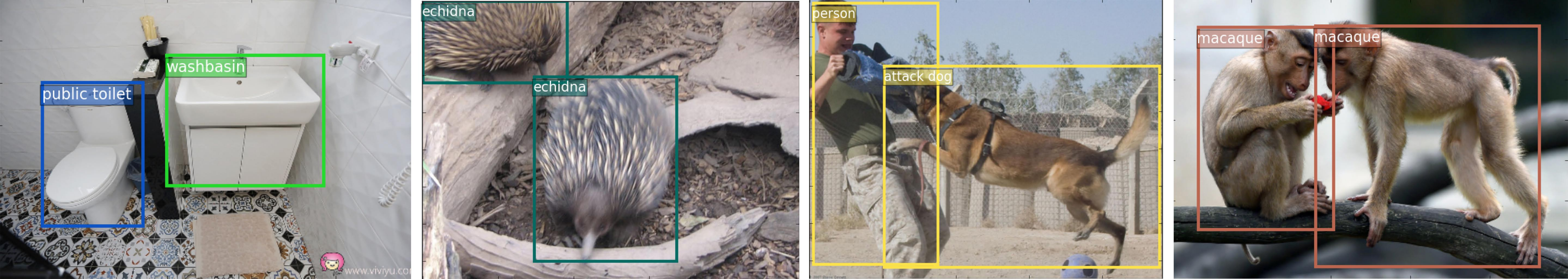}
    \caption{Detections for classes in the ImageNet3K dataset which are typically not found in common object detection datasets are shown.}
    \label{fig:3k}
\end{figure*}

We present results as we increase the number of classes on the ImageNet detection test set which contains 194 classes in Fig. \ref{fig:numc} \textcolor{red}{(c)}. In this experiment, we only use one cluster, hence the position sensitive RoI filters only predict objectness and perform bounding-box regression. The drop in performance typically reduces as we increase the number of classes. For example, there is a drop of 2\% as the number of classes is increased from 200 to 500, but from 1000 to 2000, the performance drop is only 0.3\%. Even with 3,000 classes, we obtain an mAP of 34.9\% which is 15\% better than YOLO-9000 which obtains an mAP of 19.9\%. Performance of YOLO-9000 drops to 16\% when it is evaluated on classes which are not part of the detection set (these are majority of the classes which it detects). Therefore, we perform better by 19\% on classes which are not part of the detection set compared to YOLO-9000. Although we detect 3,000 instead of 9,000 classes, our performance is more than 2 times better than YOLO-9000. Qualitative results for the R-FCN-3000 detector are also shown in Fig. \ref{fig:3k}.


To assess the effect of the number of super-classes on performance, we varied the number of super-classes and report the results. All results use a single-scale inference. Fig. \ref{fig:numc} \textcolor{red}{(a)} reports mAP for training/testing on 194 classes from the ImageNet detection dataset and Fig. \ref{fig:numc} \textcolor{red}{(b)} reports mAP for the same 194 classes while training with 1,000 object classes. The drop in performance is merely 1.7\% for a detector with only one super-class as compared to 100 super-classes for 194 class training. More interestingly, as the number of training classes are increased to 1,000, the drop is only 1.3\%, which is counter-intuitive because one would expect that using more super-classes would be helpful as we increase the number of object classes. In light of these observations, we can conclude that more crucial to R-FCN is learning an {\em objectness} measure instead of class-specific objectness.

\subsection{Are Possition-Sensitive Filters Per Class Necessary?}
To further verify our claim that detection can be modelled as a product of objectness and classification probability, we conduct more experiments on the PASCAL VOC dataset. We train a deformable R-FCN detector, as the baseline, with a ResNet-50 backbone that uses deformable position sensitive filters and obtains an mAP of 79.5\%. After training a decoupled network which predicts objectness and performs classification on RoIs, we observe a similar pattern even on this dataset. At a 0.5 overlap, the performance only drops by 1.9\% and at 0.7 by 1.5\%, Table \ref{tab:pascal_cluster}. This confirms that our proposed design changes to R-FCN are effective and only marginally deteriorate the mAP of the detector. We show a few visual examples of objectness and classification scores predicted by our class-agnostic detector in Fig \ref{fig:pascal}.

Based on these results, we explore some other alternatives of R-FCN for estimating objectness. First, we just use RPN scores as the objectness measure and classify the proposals with our network (which is a linear classifier). Then, we add a bounding box regression step on the proposals, as they are already class agnostic. These two baselines are significantly worse than R-FCN. The mAP of only RPN is very poor at an overlap of 0.7. Although bounding-box regression provides a boost of 35\% at 0.7 overlap, the performance is still 15\% worse than R-FCN. Since RPN uses an overlap of 0.7 for assigning positives and 0.3 for assigning negatives, we decided to change these two thresholds to 0.5 and 0.4 respectively, like \cite{lin2017focal}. We train two versions of RPN, on conv4 and conv5 and present the results. These results show that performance with RPN also improves after changing the overlap criterion and with better features, so other objectness measures could also be an alternative for R-FCN. Results for these experiments are presented in Table \ref{tab:pascal_rpn}.

 \begin{figure*}[ht!]
    \center
    \includegraphics[width=1.00\linewidth]{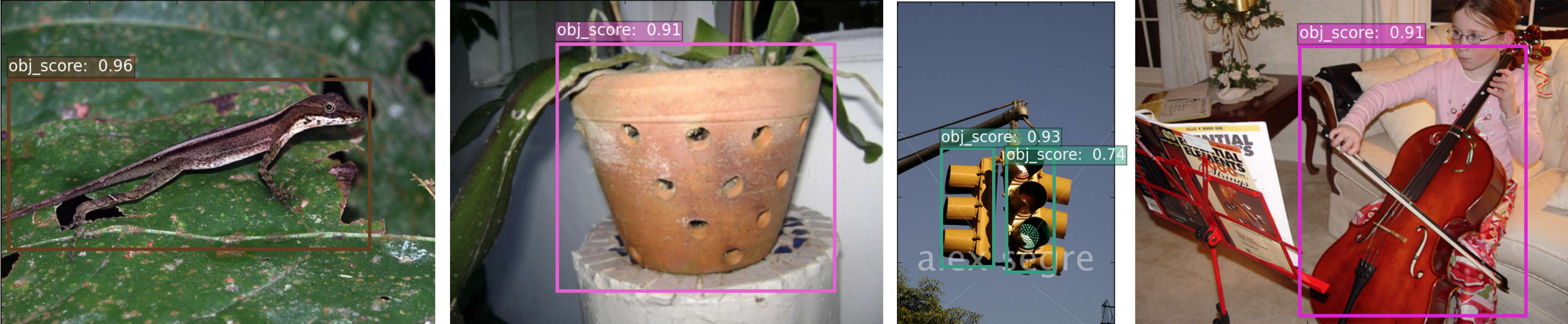}
    \caption{Objectness scores on images containing unseen object-classes from the ImageNet detection dataset. 
    }
    \label{fig:20C}
\end{figure*}



\begin{table}[t!]
    \centering
    \begin{tabular}{c|c|c|c|c|c|c}
        \hline
          Ov$^{0.5}$ & Ov$^{0.7}$ & C4 & C5 & BBR & mAP$^{0.5}$ & mAP$^{0.7}$ \\
        \hline
        $\times$ & \checkmark & \checkmark & $\times$ & $\times$ & 47.3 & 12.7 \\
        $\times$ & \checkmark & \checkmark & $\times$ & \checkmark & 65.1 & 47.8 \\
        \checkmark & $\times$ & \checkmark & $\times$ & $\times$ & 52.0 & 16.0 \\
        \checkmark & $\times$ & \checkmark & $\times$ & \checkmark & 66.8 & 49.7 \\
        \checkmark & $\times$ & $\times$ & \checkmark & $\times$ & 70.6 & 44.1 \\
        \checkmark & $\times$ & $\times$ & \checkmark & \checkmark & 74.1 & 56.9 \\
        \hline
    \end{tabular}
    \newline
    \caption{Results for different versions of RPN scores used for objectness are reported. C4 and C5 denote if RPN is applied on Conv4 or Conv5 feature-map. Ov$^{0.5}$, Ov$^{0.7}$ denotes if the overlap for assigning positives in RPN is 0.5 or 0.7. BBR denotes if bounding box regression of deformable R-FCN is used or not.}
    \label{tab:pascal_rpn}
\end{table}

\begin{table}[t!]
    \centering
    \begin{tabular}{c|c|c}
        \hline
          Method & mAP$^{0.5}$ & mAP$^{0.7}$ \\
        \hline
         D-R-FCN (decoupled)  & 77.6 & 63.8\\
        \hline
         D-R-FCN & 79.5 & 65.3 \\
        \hline
    \end{tabular}
    \newline
    \caption{Results of D-R-FCN (ResNet-50) and our decoupled version where the R-FCN classification branch only predicts objectness.}
    \label{tab:pascal_cluster}
\end{table}

\begin{figure}[t]
    \center
    \includegraphics[width=0.99\linewidth]{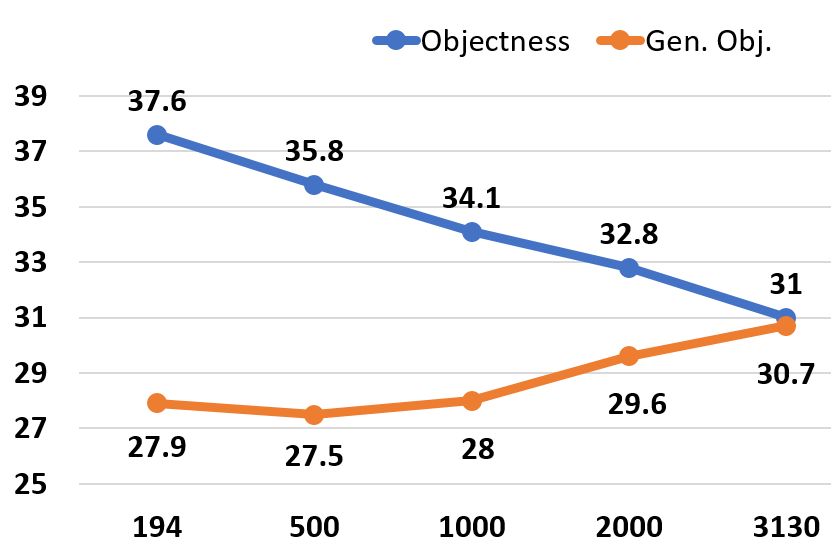}
    \caption{The mAP scores on a held out set of 20 classes for Generalized Objectness and Objectness baseline.}
    \label{fig:gen}
\end{figure}

\subsection{Generalization of Objectness on Unseen Classes}
\begin{figure*}[t]
    \center
    \includegraphics[width=0.91\linewidth]{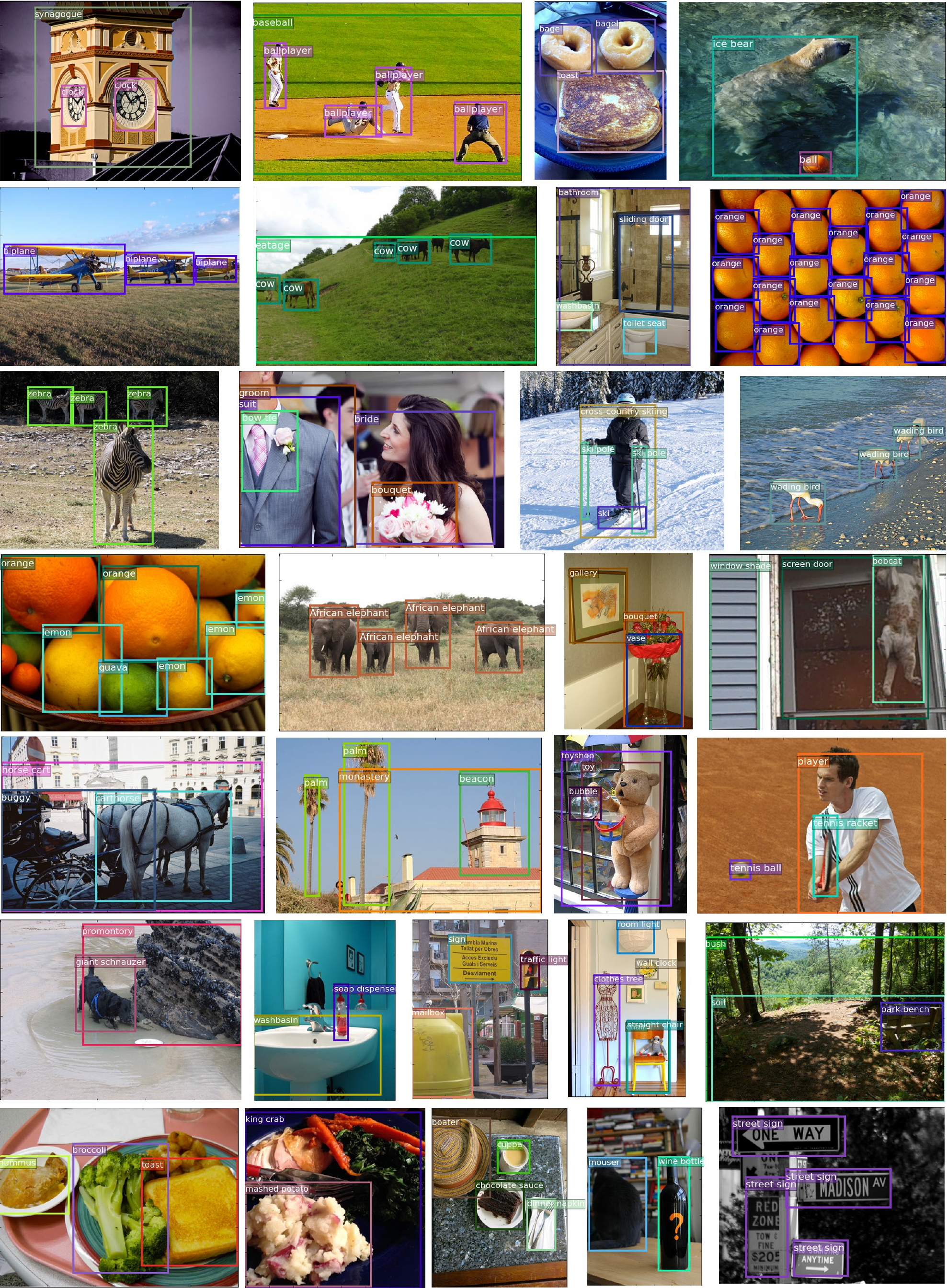}
    \caption{Qualitative results on COCO. We miss some objects, but the figure depicts the diversity of classes detected by R-FCN-3000!}
    \label{fig:coco}
\end{figure*}

We evaluate the generalization performance of our objectness detector on a held out set of 20 classes. In this experimental setting, we train two objectness detectors - {\bf OB} (objectness baseline), which includes the 20 object classes during training and  {\bf GO }(generalized objectness), which does not. For both the settings, the same classifier is used with different objectness detectors. OB and the classifier are trained on 194, 500, 1000, 2000 and 3130 classes and GO on 174, 480, 980, 1980 and 3110 classes. While going from 194 to 500 classes, the number of classes increase significantly but the number of samples do not (see Table \ref{tab:stats}); therefore, the mAP of OB drops by 1.8\%. Since more samples help in improving the objectness measure for GO, the performance drop is only marginal (Fig \ref{fig:gen}). Increasing the number of classes to 1000 and 2000 improves the mAP of GO, implying that the improvement in objectness can overshadow the performance drop caused by increasing the number of classes. Fig \ref{fig:gen} clearly shows that the initial gap of 9.7\% in the performance drops to 0.3\% as the number of classes increase. We also compared {\bf OB} with {\bf GO} when we remove all the 194 classes in ImageNet detection set and present the results in Table \ref{tab:got}. Note that the performance drop is 3\% even after removing 10\% of the instances in the dataset (all of which belong to the classes in the test set). It strongly indicates that objectness learned on thousands of classes generalizes to novel unseen classes as well. A few qualitative results for such cases are shown in Fig. \ref{fig:20C}. We also show some qualitative results on a few COCO images in Fig \ref{fig:coco}. Note that we did not train on any detection images!

\begin{table}[t!]
    \centering
    \begin{tabular}{c|c|c}
        \hline
         Classes  & Objectness & Generalized Objectness \\
        \hline
         20 & 31 & 30.7 \\
        \hline
         194 & 34.9 & 32 \\
        \hline
    \end{tabular}
    \newline
    \caption{mAP of Objectness and Generalized Objectness on held out classes in the ImageNet detection set.}
    \label{tab:got}
\end{table}

\section{Conclusion and Future Directions}

We demonstrate that it is possible to predict a universal objectness score by using only one set of filters for object vs. background detection. This objectness score can simply be multiplied with the classification score for detecting objects with only a marginal drop in performance. Finally, we show that the objectness learned generalizes to unseen classes and the performance increases with the number of training object classes. It bolsters the hypothesis of the universality of objectness.

This paper presents significant improvements for large-scale object detection but many questions still remain unanswered. Some promising research questions are - How can we accelerate the classification stage of R-FCN-3000 for detecting 100,000 classes? A typical image contains a limited number object categories - how to use this prior to accelerate inference? What changes are needed in this architecture if we also need to detect objects and their parts? Since it is expensive to label each object instance with all valid classes in every image, can we learn robust object detectors if some objects are not labelled in the dataset?


{\small
\bibliographystyle{ieee}
\bibliography{egbib}
}

\end{document}